\documentclass{article}
\usepackage{spconf,amsmath,epsfig}

\usepackage{url}
\usepackage{times}
\usepackage{graphicx}
\usepackage{amssymb}
\usepackage{booktabs}
\usepackage{caption}
\usepackage{subcaption}
\usepackage{bbm}

\graphicspath{ {images/} }

\usepackage{xcolor}

\begin{document}\sloppy

\def\x{{\mathbf x}}
\def\L{{\cal L}}

\title{
Correction by Projection: Denoising Images \\
with Generative Adversarial Networks 
}
%

\threeauthors{Subarna Tripathi}{UC San Diego}{Zachary C. Lipton}{UC San Diego}{Truong Q. Nguyen} {UC San Diego}

\maketitle

\begin{abstract}
Generative adversarial networks (GANs)
transform low-dimensional latent vectors into visually plausible images. 
If the real dataset contains only clean images, then ostensibly, the manifold learned by the GAN should contain only clean images. 
\textbf{In this paper,} we propose to denoise corrupted images by finding the nearest point  on the GAN manifold,
recovering latent vectors by minimizing distances in image space. 
We first demonstrate that given a corrupted version of an image that truly lies on the GAN manifold,
we can approximately recover the latent vector and denoise the image,
obtaining significantly higher quality, comparing with BM3D.
Next, we demonstrate that latent vectors recovered from noisy images exhibit a consistent bias. 
By subtracting this bias before projecting back to image space, 
we improve denoising results even further. 
Finally, even for unseen images, our method performs better at denoising better than BM3D. 
Notably, the basic version of our method
(without bias correction) 
requires no prior knowledge on the noise variance. 
To achieve the highest possible denoising quality, the best performing signal processing based methods, such as BM3D, 
require an estimate of the blur kernel. 
\end{abstract}

\section{Introduction} \label{introduction}
Generative adversarial networks (GANs) \cite{goodfellow2014generative,radford2015unsupervised} 
exploit the discriminative power of deep neural networks for the task of generative modeling. 
A GAN consists of two models: a \emph{generator} and a \emph{discriminator}.
The generator maps samples from a low-dimensional latent space onto the space of images. 
The discriminator tries to distinguish between images produced by the generator and real images.
To coerce the generator to produce images that match the distribution of real images, 
we optimize it to fool the discriminator. 
It has been shown that various GAN minimax objectives are equivalent to minimizing corresponding divergences between the real and generated data.

\begin{figure}
\begin{center}
	\includegraphics[width=1\linewidth]{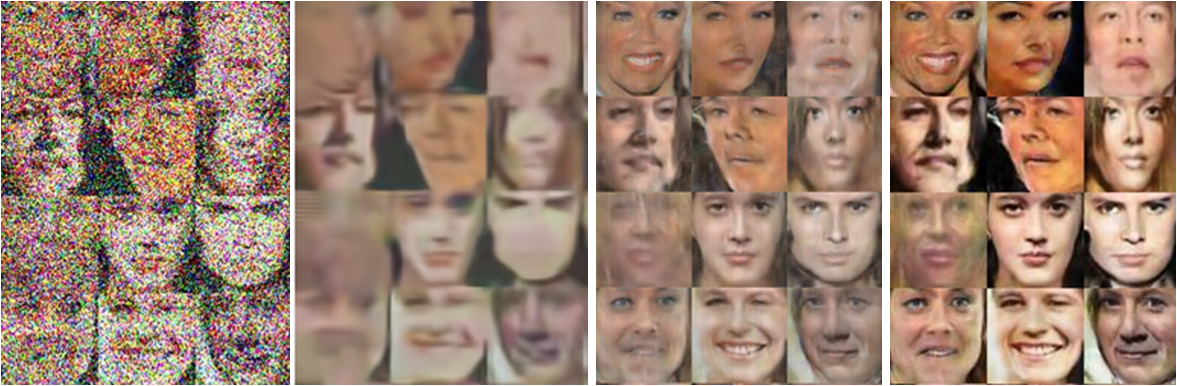} 
\end{center}
   \caption{Images with Gaussian noise (left), denoised images with BM3D (second), denoised images with proposed method(third), and the ground truth images (right) }
\label{intro}
\end{figure}

 Over the last several years, 
many researchers have successfully applied variants of GANs~\cite{GAN_zoo_17} 
to tasks including synthesizing 2D images, face processing, image completion, image editing, 3D objects generation and reconstruction, fashion image generation, image and video super-resolutions, video generation, image-to-image translation, text-to-image generation, audio and text synthesis.

While generating images can be useful,
we often want to infer latent representations given images.
As an example, the latent code recovery methods is also used for generating fashion images that have horizontal symmetry ~\cite{Makkapati2017}. 
It is known that vectors that are close in latent space, generate visually similar images.
Algebraic operations in latent vector space often lead to meaningful
corresponding operations in image space. 

However, the original GAN formulation gives no out-of-the-box method to reverse the mapping, 
projecting images back into latent space. 
How best to perform the 
reverse mapping (from image space to latent space) 
remains an open research problem.
Authors in \cite{donahue2016adversarial} suggests an extension to GAN in which a third model explicitly learns the reverse mapping.
\cite{creswell2016inverting} suggest that inverting the generator is difficult, noting that, in principle, a single image $\phi({\boldsymbol{z}})$ may map to multiple latent vectors $\boldsymbol{z}$. 
They propose a gradient-based approach to recover latent vectors
and evaluate the process on the reconstruction error 
in image space. 

Recently,~\cite{LiptonT17} proposes to recover latent vectors using a gradient-based method using ``stochastic clipping'', and achieve successful recovery $100\%$ of time given a certain residual threshold. 
The idea of ``stochastic clipping'' is based on the notion that the latent vectors have close to zero probability of landing on the boundary values.  

\textbf{In this paper,} we show that latent code recovery can be used to denoise and deblur images, 
building on the method of ~\cite{LiptonT17}.
First, we show that for corrupted versions of images that are actually generated by our trained GAN, 
we can significantly denoise them, achieving  higher quality as measured by PSNR, 
compared to 
BM3D~\cite{BM3D},
one of the best signal processing-based denoising methods. 
Next, we demonstrate 
that \emph{deblurring} can be treated as an attribute in latent space, 
given the noise variance. 
Arithmetic on \emph{deblurring} latent vector space denoises and deblurs the images even further. 
Finally, we show that even for unseen images, our method appears to denoise better than BM3D (Fig~\ref{intro}).
To our knowledge, this is the first empirical demonstration 
that recovery of latent vectors in DCGAN can be used for image denoising and deblurring.
After adding even significant amounts of Gaussian noise to images,
we denoise the images with higher fidelity comparing with state-of-the-art denoising method. 


\section{Related Work} \label{related_work}

Image denoising is an actively researched inverse problem in low-level vision for last couple of decades. 
It has rich literature in traditional signal processing based methods. 
See~\cite{image_filtering_survey_13} for a detailed survey. BM3D~\cite{BM3D} is one of the best methods on image denoising in that domain. 
Recently, many researchers have proposed discriminative learning based methods for image denoising.
Typically, these methods learn the image prior models and corresponding denoising function using CNNs. 
See
~\cite{cnn_denoising_FDD_net_17, cnn_denosing_ZhangZCM016, cnn_denoising_17}
for overviews on these methods. 

In this paper, 
we focus on the use of GANs for denoising and deblurring.
Specifically, we project noisy images onto the range of the GAN by attempting to recover the latent vector which corresponds to the closest point on the GAN manifold. 
Several papers attempt gradient-based methods for inverting deep neural networks.
Authors in~\cite{mahendran2015understanding} invert discriminative CNNs for the purpose of interpreting hidden representations.
To invert generative models, \cite{creswell2016inverting} and \cite{metz2016unrolled} both optimize over latent vectors to minimize distance in image space, but neither reported that the inference procedure could faithfully recover ground truth latent vectors or that the inferred vectors (across multiple runs) tended to be proximal in latent space. 
In a different approach, ~\cite{donahue2016adversarial}, and \cite{dumoulin2016adversarially},
learn separate neural network encoders 
for performing the reverse mapping.
The latent code recovery methods \cite{LiptonT17, compressed_sensing2017} is also used for generating fashion images that have horizontal symmetry ~\cite{Makkapati2017}. 
Authors in ~\cite{LiptonT17} also note that the reverse projection can be used to remove Gaussian noise. 
However, they do not investigate any potential for latent vector arithmetic for deblurring. 
Authors in~\cite{Manipulate_ip_spaces_17} allow the model to change the input vector that leads to better images according to the discriminator.

\section{Method} \label{method}

Recovery from an image generated by the generator, and from a real image are different. 
Former one can be considered as an inverse operation of an existing forward operation. 
However, attempting to recover the latent vector from a real image is more like a projection of it onto the manifold learned by the generator. 

Our approach of recovering the latent vectors is based on \cite{LiptonT17}.
In order to recover the latent vector of a generated image, 
we produce an image $\phi(\boldsymbol{z})$ for a 
latent vector $\boldsymbol{z}$. 
We then initialize a new, random vector $\boldsymbol{z'}$ 
of the same shape as $\boldsymbol{z}$.
This new vector $\boldsymbol{z'}$ 
maps to a corresponding image $\phi(\boldsymbol{z'})$.
In order to reverse engineer the input $\boldsymbol{z}$, 
we successively update the components of $\boldsymbol{z'}$ in order to push the representation $\phi(\boldsymbol{z'})$ closer to the original image $\phi(\boldsymbol{z})$.
In our experiments we minimize the $L_2$ norm, yielding the following optimization problem:
$$\min_{\boldsymbol{z'}} || \phi(\boldsymbol{z}) - \phi(\boldsymbol{z'})||^2_2.$$
We optimize over $\boldsymbol{z'}$ by gradient descent, performing the update $\boldsymbol{z'} \gets \boldsymbol{z'} - \eta \nabla_{\boldsymbol{z'}} || \phi(\boldsymbol{z}) - \phi(\boldsymbol{z'})||^2_2 $ until some convergence criteria is met. Only solving for this optimization problem refers to as \emph{no clipping}.

All latent vectors are sampled uniformly from the $[-1,1]^{100}$ hyper-cube. 
To enforce this constraint, 
we apply the modified optimization
$$\boldsymbol{z'} \gets \text{clip}(\boldsymbol{z'} - \alpha \nabla_{\boldsymbol{z}'} || \phi(\boldsymbol{z}) - \phi(\boldsymbol{z'})||^2_2 ).$$
For \emph{projected gradient},
we replace components that are too large 
with the maximum allowed value 
and components that are too small 
with the minimum allowed value.
The authors in \cite{LiptonT17} 
introduce a heuristic technique called \emph{stochastic clipping}.
When using stochastic clipping, instead of setting components to $-1$ or $1$, we reassign the clipped components uniformly at random in the allowed range. 

\begin{figure}
\begin{center}
	\includegraphics[width=1\linewidth]{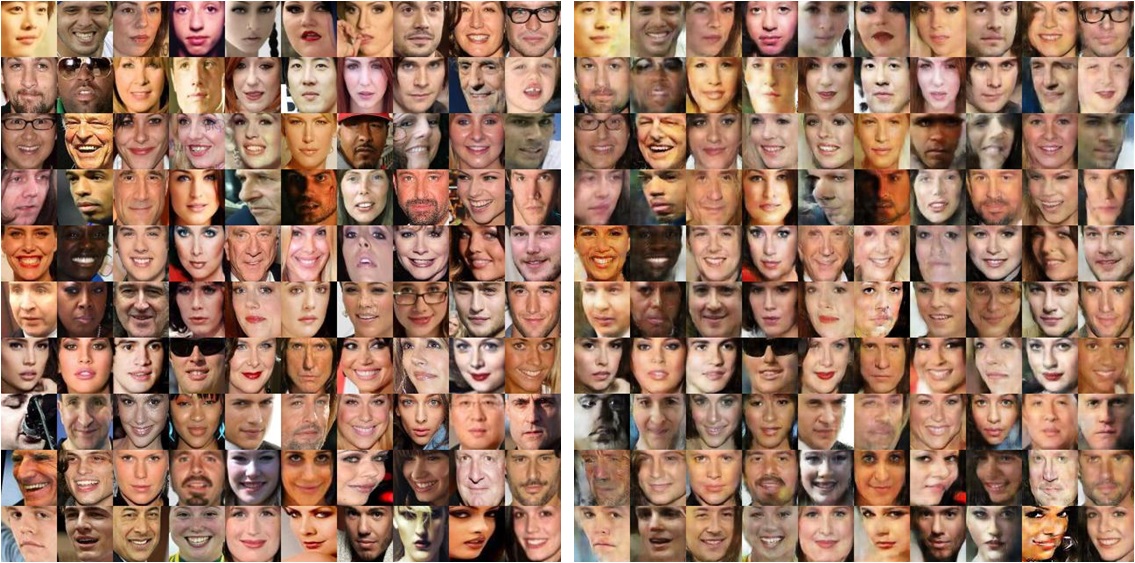} 
\end{center}
   \caption{Example recovered images from CelebA dataset (no added noise)  with stochastic gradient~\cite{LiptonT17}. }
\label{CelebA_recovery}
\end{figure}

\begin{figure}
\begin{center}
	\includegraphics[width=1\linewidth]{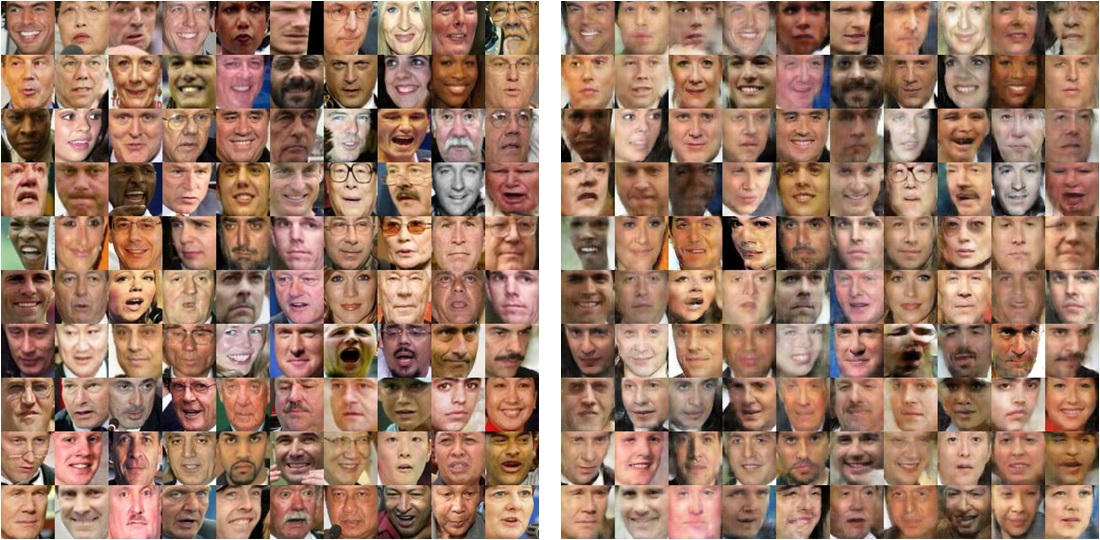} 
\end{center}
   \caption{Example recovered images from LFW dataset (no added noise) with stochastic gradient~\cite{LiptonT17}. }
\label{LFW_recovery}
\end{figure}

For the recovery of a real image the forward mapping $\phi(\boldsymbol{z})$ does not necessarily exist. We perform the following optimization in this case. $$\min_{\boldsymbol{z'}} || I - \phi(\boldsymbol{z'})||^2_2.$$ with \emph{stochastic clipping}, where $I$ denotes real images.
We refer this latent vectors recovery methods as LVR.
The generator in our experiments is trained with CelebA dataset~\cite{celebA_liu2015}. 
Example real images from CelebA dataset and their recovery using the LVR are shown in Fig~\ref{CelebA_recovery}. 
Additionally, some images from  
LFW~\cite{LFWTech} (the generator never saw these images) and their recovery are shown in Fig~\ref{LFW_recovery}.

Authors in ~\cite{LiptonT17} note that the reverse projection can be used to remove Gaussian noise. In order to explore the denoising potential within this framework, we corrupt the real images with varying levels of Gaussian noise variance, and apply LVR for $\min_{\boldsymbol{z'}} || (I+\eta) - \phi(\boldsymbol{z'})||^2_2$, where $I$ denotes real images and $\eta$ denotes Gaussian noise variance. The above denoising methods do not require the noise variance a priori unlike traditional denoising methods, and still produce better denoising results.

In case of high noise variance, the recovered images using LVR methods appear to be blurred. 
In order to explore the deblurring potential, we apply LVR on $N$ generated images with $K$ different noise variances represented by ($\phi(\boldsymbol{z_i})+\eta_k$) and recover the corresponding latent vectors $\boldsymbol{z'_i}_k$ for those noise levels. Here, $i$ indexes over $1$ to $N$ and $k$ indexes over $1$ to $K$. $z_k$ is the difference between average over $N$ sample $z$ and the average of their corresponding recovered $z'_k$ for noise level $k$. 
We observe that, adding $z'_k$ to recovered latent vectors from other  generated images with noise variance $k$ sharpens the recovered images. Interestingly, this happens while denoising real images. 
Empirically, there seems to exist vectors in latent space that can add sharpness in image space. 
If we denote the sharpness attribute in latent vector space as ${z_{sharp}}_k$ for noise variance $k$, interestingly, 
we find the quality of $\phi(\boldsymbol{z'}+{z_{sharp}}_k)$ is higher than $\phi(\boldsymbol{z'})$ for recovered latent vectors $z'$ with LVR. 



\section{Experimental Results} \label{results}

We now summarize our experimental findings. All experiments are conducted with DCGANs as described by \cite{radford2015unsupervised} and re-implemented in Tensorflow by \cite{amos2016image}. We made  necessary changes on top of \cite{amos2016image}.
We train the generator using the CelebA~\cite{celebA_liu2015} dataset.

\begin{figure*}
\begin{center}
	\includegraphics[width=1\linewidth]{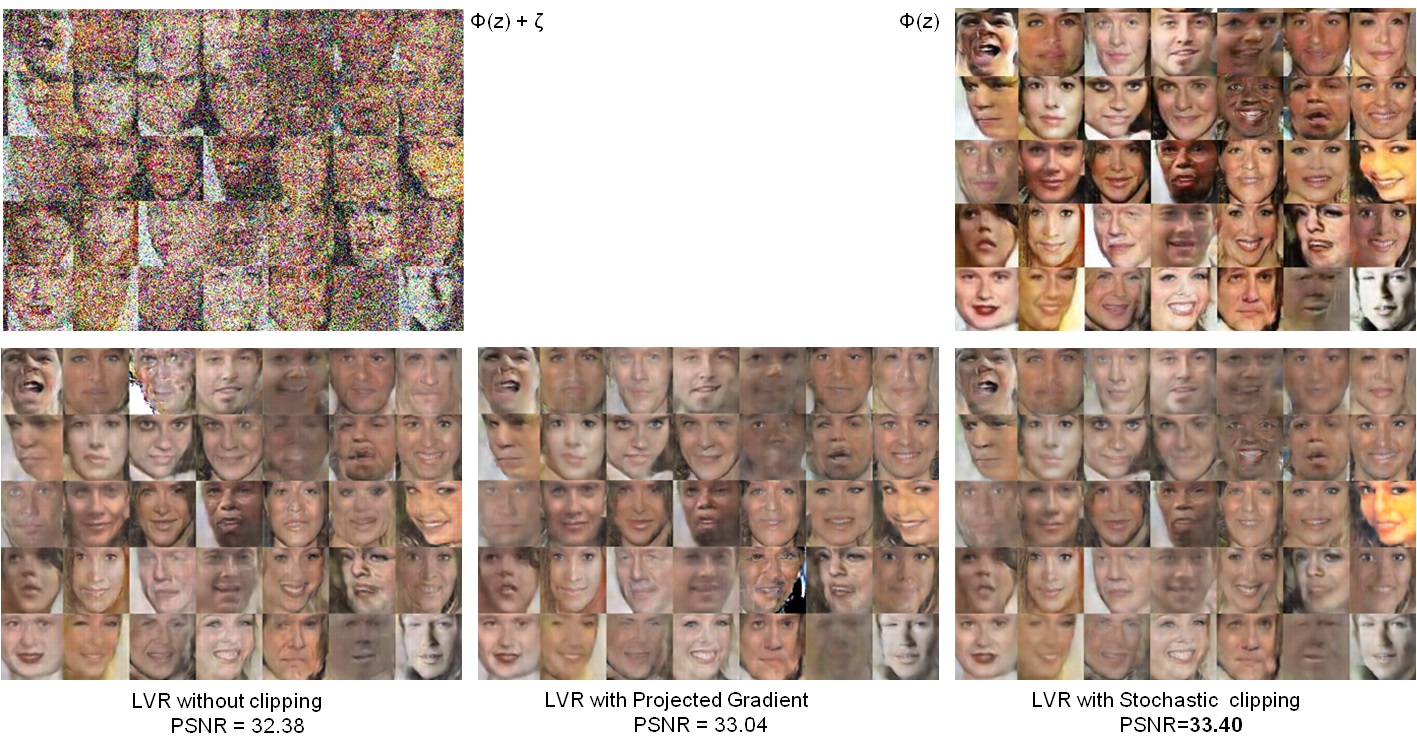} 
\end{center}
   \caption{Image De-noising Results with LVR methods~\cite{LiptonT17}. Top row: input noisy images to the algorithm; Reference generated images. Bottom row: Recovered images by LVR with \emph{no clipping}; \emph{projected gradient} and \emph{stochastic gradient} methods. }
\label{denoising-LVR}
\end{figure*}

\begin{table}[hbt!]
\centering
\begin{tabular}{lll} 
& \multicolumn{2}{c}{\textbf{PSNR of reconstructed images with~\cite{LiptonT17}}}
\\
\toprule
No clipping 
& Projected Gradient & Stochastic Clipping 
\\
\midrule
$32.38$ & $33.04$ &$33.40$ 
\\
\bottomrule
\end{tabular} 
\caption{
Reconstructed image quality evaluation for \emph{No clipping}, \emph{projected gradient} and \emph{stochastic gradient} strategies as described in~\cite{LiptonT17}. 
}
\label{tab:reconstruction-results}
\end{table}

We first perform the experiments for denoising of generated images 
after adding Gaussian noise with variance of $127$ pixels units
by applying different strategies as described in~\cite{LiptonT17}. 
\emph{Stochastic clipping} method outperforms \emph{projected gradient} which performs better than \emph{no clipping} strategy. Stochastic clipping method obtains more than $1$dB higher PSNR than the without clipping strategy. Figure~\ref{denoising-LVR} shows the corresponding visual results. 

We notice that the denoised images are significantly smoother than the images with which the generator was trained on. In order to improve the denoising results further, we explore the latent space for additional sharpness. 
We compare the latent space sharpness attribute-based method as \emph{LVR-SA}, and always use the baseline of latent vector recovery with stochastic clipping. 

\begin{figure}
\begin{center}
	\includegraphics[width=.49\linewidth]{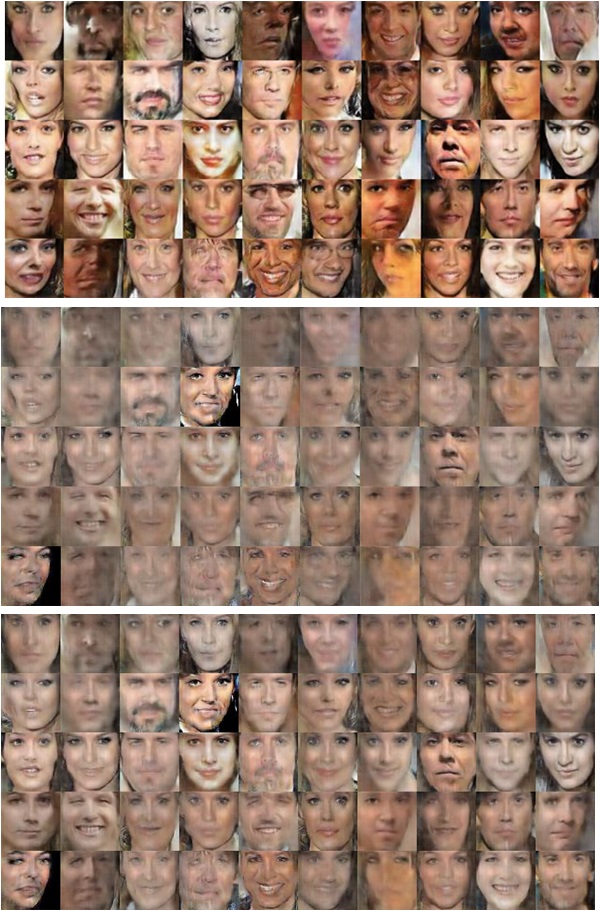} 
    \includegraphics[width=.49\linewidth]{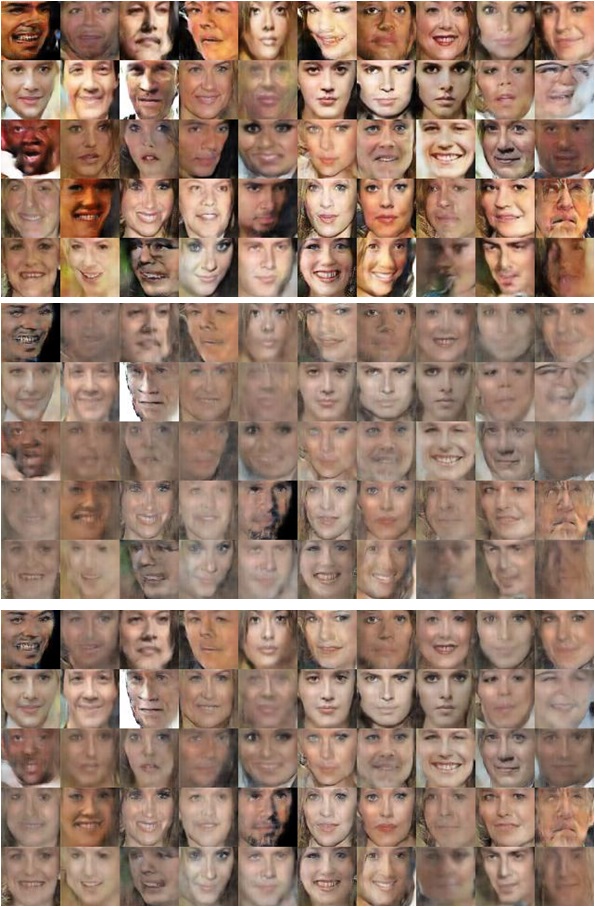}
\end{center}
   \caption{Sharpness attribute heuristics for images from from celebA dataset (left) and LFW dataset (right). The sharpness attribute $S$ is added to the two scenarios. }
\label{sharpness_set}
\end{figure}

As described in section~\ref{method}, we first generate a set of $400$ images, $I_g$, corrupt them with Gaussian noise with different variance. For each noise variance, we recover the latent vectors using stochastic clipping strategy. We then take the difference between average latent vectors that generated the image and the average 
latent vectors that were recovered. This difference serves as sharpness attribute. 
First, we add this sharpness attribute to the recovered vectors for those $I_g$ images. 
We observe that after adding the sharpness attribute, the PSNR increases upto $1$ dB.  
Next, we apply the same sharpness attribute while recovering a different set of real images, $I_r$. 
Interestingly, the same sharpness attribute helps increase the PSNR by more than $0.5$ dB. 
Table~\ref{tab:sharpness attribute} summarizes the above results, where the first row corresponds to the former case and the second row corresponds to the later scenario respectively. Figure~\ref{sharpness_set} shows the corresponding visualizations. 
For each column, the top images are the original images. We add Gaussian noise to those images and recover them in the middle row with stochastic clipping. Third row denotes the recovered images after adding the sharpness attribute $SA$. 
The generator is trained only on celebA~\cite{celebA_liu2015}. In the left column, the images are sampled from CelebA dataset and in the right, the images are sampled from LFW~\cite{LFWTech} dataset. 

\begin{table}[hbt!]
\centering
\begin{tabular}{llll} 
& \multicolumn{3}{c}{\textbf{Effect of Sharpness attribute}}
\\
\toprule
%
& LVR & LVR-SA
\\
\midrule
same image sets & $17.1$ &$18.01$ \\
different image sets & $18.43$ &$19.02$ 
\\
\bottomrule
\end{tabular} 
\caption{Adding sharpness attribute to the recovered vectors increases enhances the image quality.
}
\label{tab:sharpness attribute}
\end{table}



\begin{table}[hbt!]
\centering
\begin{tabular}{llll} 
& \multicolumn{3}{c}{\textbf{Methods}}
\\
\toprule
$\mathbf{\sigma}$ (in pixels)
& BM3D & LVR & LVR-SA
\\
\midrule
$127$ & $29.10$ &$33.4$ & $33.60$\\
$184$ & $18.55$ &$22.0$ & $22.21$
\\
\bottomrule
\end{tabular} 
\caption{
Reconstructed image quality evaluation by PSNR for different noise variance. For best BM3D reconstruction, corresponding noise variance is given. Unlike LVR, Denoising by BM3D and the proposed LVR-SA require noise variance. 
}
\label{tab:reconstruction-results}
\end{table}

\begin{figure*}
\begin{center}
	\includegraphics[width=1\linewidth]{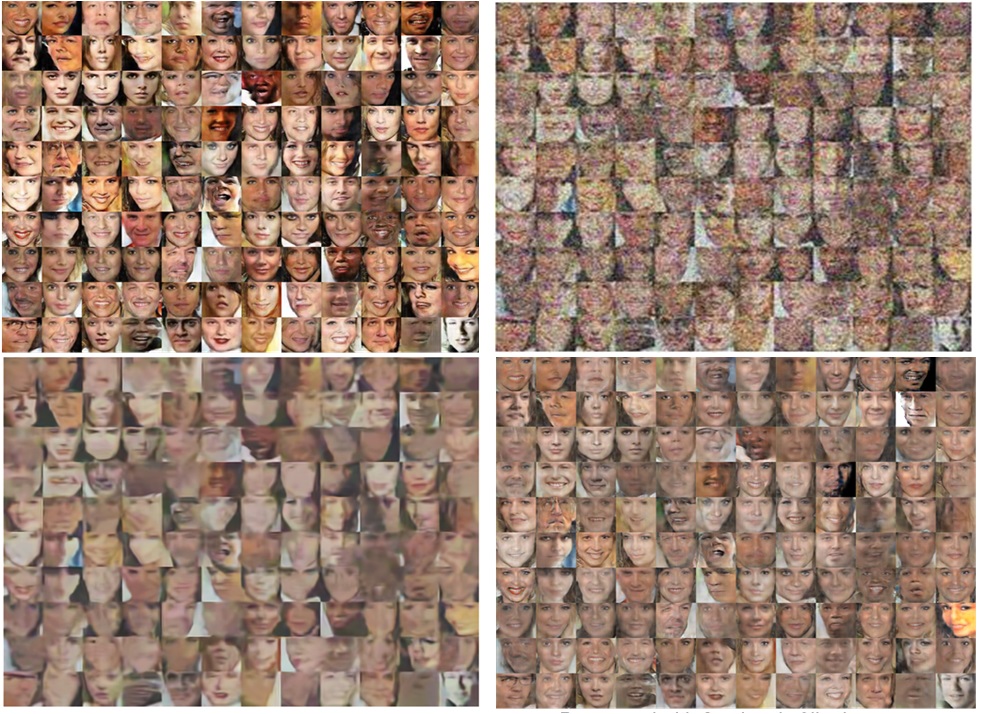} 
\end{center}
   \caption{Image denoising Results. Top row: original images from celebA dataset, input images to the algorithm with added Gaussian noise ($127$ noise variance in pixels) on input images.  Bottom row: denoised images with BM3D, denoised images with proposed method. }
\label{denoising-results}
\end{figure*}

Finally we compare the denoising results 
utilizing the above sharpness attribute method 
with stochastic clipping LVR and BM3D. Table~\ref{tab:reconstruction-results} shows the results for different noise variances. 
The RGB images use $8$-bit representation. The pixels can take values in the range of $[0, 255]$ both inclusive for R,G, and B space.
We experiment with high noise variance values. 
For low noise variance, the recovered images with different methods yield similar quality image deniosing. 
Latent vector recovery (LVR) with stochastic clipping yields better denoising results. 
This LVR method does not require the noise variance as a prior knowledge.  
Moreover, adding the sharpness latent vector (LVR-SA) improves the image quality on top of LVR. 
BM3D and LVR-SA both require a-priori knowledge of the noise variance. 
Figure~\ref{denoising-results} demonstrates the corresponding visual results. Denoising with the proposed method recovers clearly better quality images than BM3D.

\section{Conclusions} \label{conclusions}

We show that latent vector recovery from GAN, in practice, can be used for image denoising. 
The denoised image quality is superior to methods such as BM3D. 
Additionally, we show that sharpness can be treated as an attribute in latent vector space. Adding this sharpness attribute leads to even higher quality image denoising than simple latent vector recovery based methods.

\bibliographystyle{IEEEbib}
\bibliography{icme2018template}

\end{document}